\documentclass[]{opendatalab}
\usepackage{xcolor}
\usepackage{longtable}
\usepackage{listings}
\usepackage[numbers]{natbib}

\usepackage{enumitem}
\usepackage[table]{xcolor}
\usepackage{pifont} 
\usepackage{fontawesome5}

\usepackage{listings}
\usepackage{xcolor}
\newcommand{\faHuggingFace}{%
  \raisebox{-0.15em}{%
    \includegraphics[height=1em,keepaspectratio]{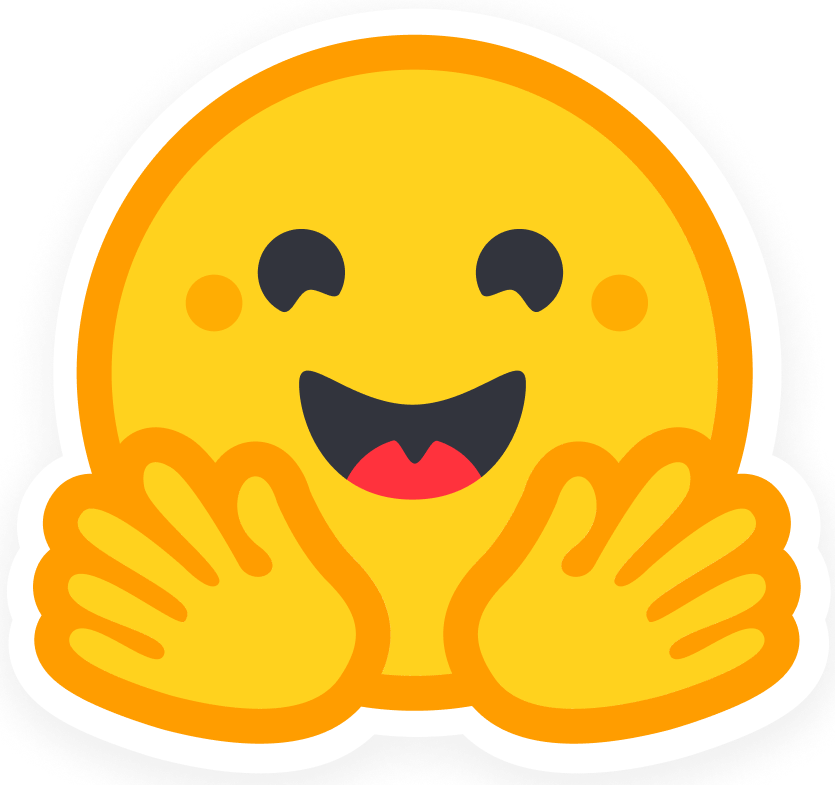}%
  }%
}

\newcommand{\faModelscope}{%
  \raisebox{-0.15em}{%
    \includegraphics[height=1em,keepaspectratio]{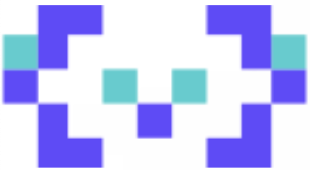}
  }%
}

\definecolor{codegreen}{rgb}{0,0.6,0}
\definecolor{codegray}{rgb}{0.5,0.5,0.5}
\definecolor{codepurple}{rgb}{0.58,0,0.82}
\definecolor{backcolour}{rgb}{0.95,0.95,0.92}
\definecolor{promptcolor}{HTML}{D1D0F2}
\definecolor{promptcolorheader}{HTML}{bdbcec}
\newcommand{\promptbox}[2]{
\begin{tcolorbox}[
top=0.3em,bottom=0.3em,left=0.5em,right=0.5em,
toptitle=0.3em,bottomtitle=0.2em,boxsep=0pt,
colframe=promptcolorheader,colback=promptcolor!50,boxrule=0.5pt,
]
\footnotesize
\end{tcolorbox}
}
\lstdefinestyle{mystyle}{
    backgroundcolor=\color{backcolour},   
    commentstyle=\color{codegreen},
    keywordstyle=\color{magenta},
    numberstyle=\tiny\color{codegray},
    stringstyle=\color{codepurple},
    basicstyle=\ttfamily\footnotesize,
    breakatwhitespace=false,         
    breaklines=true,                 
    captionpos=b,                    
    keepspaces=true,                 
    numbers=left,                    
    numbersep=5pt,                  
    showspaces=false,                
    showstringspaces=false,
    showtabs=false,                  
    tabsize=2
}

\title{
Unlocking Data Value in Finance: A Study on Distillation and Difficulty-Aware Training}

\author[1,2]{Chuxue Cao}
\author[1]{Honglin Lin}
\author[1]{Zhanping Zhong}
\author[1]{Xin Gao}
\author[1]{Mengzhang Cai}
\author[1]{Conghui He}
\author[2*]{Sirui Han}
\author[1*]{Lijun Wu}

\affiliation[1]{Shanghai Artificial Intelligence Laboratory, OpenDataLab, OpenDataArena}
\affiliation[2]{Hong Kong University of Science and Technology}

\abstract{

Large Language Models (LLMs) have demonstrated strong general capabilities, yet their deployment in finance remains challenging due to dense domain-specific terminology, stringent numerical reasoning requirements, and low tolerance for factual errors. We conduct a controlled empirical study showing that in specialized vertical domains, performance is largely determined by the quality and difficulty/verifiability profile of post-training data. We introduce \textbf{ODA-Fin-SFT-318k}, constructed via multi-stage distillation and verification to produce high-quality Chain-of-Thought supervision, and \textbf{ODA-Fin-RL-12k}, curated for hard-but-verifiable tasks that balance reward precision and task diversity. Using standard SFT and RL pipelines, we show that high-quality CoT distillation establishes a robust foundation during SFT, while difficulty- and verifiability-aware sampling improves RL generalization. Evaluated on nine benchmarks spanning general financial tasks, sentiment analysis, and numerical reasoning, our ODA-Fin-RL-8B consistently surpasses open-source state-of-the-art (SOTA) financial LLMs of comparable size. We release our ODA-Fin-SFT-318k and ODA-Fin-RL-12k datasets, along with trained models to advance data-centric financial AI research.

}

\date{\today}
\correspondence{Lijun Wu, \email{wulijun@pjlab.org.cn}, Sirui Han, \email{siruihan@ust.hk}}

\metadata[Data \& Model]{\url{https://huggingface.co/collections/OpenDataArena/oda-finance}}
\begin{document}

\vspace{-0.3cm}
\begin{figure}[th]
    \hspace{5cm}
    \includegraphics[width=0.3\linewidth]{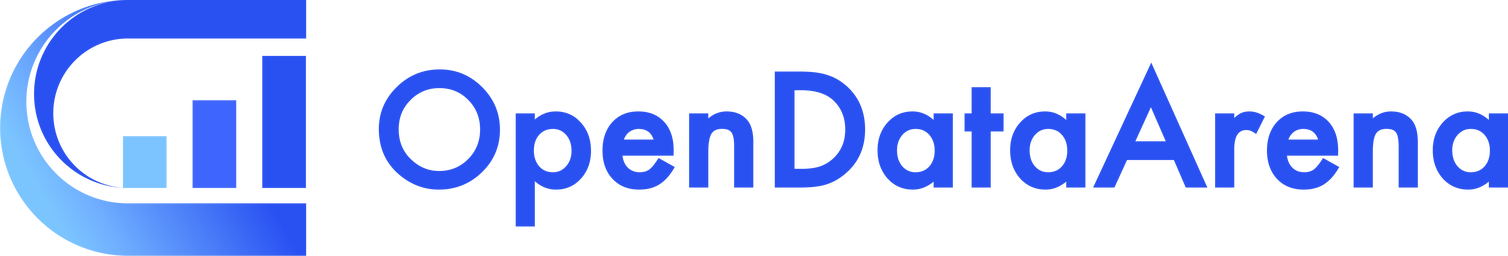}
    % \caption{Caption}
    \label{fig:logo}
\end{figure}
\vspace{-0.2cm}

\maketitle

\begin{figure}[htbp]
\centering
  \includegraphics[width=\textwidth]{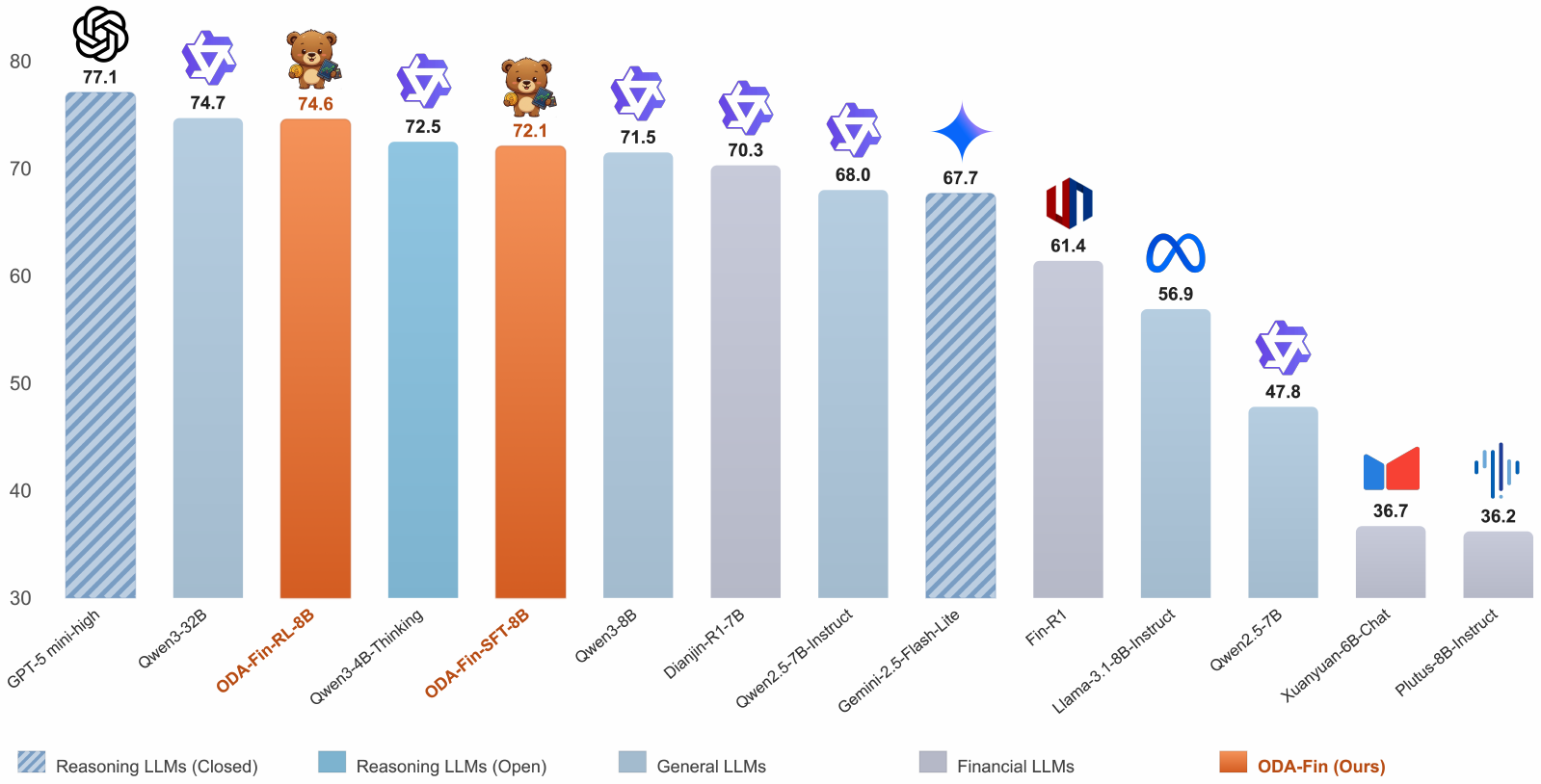}
  \caption{Average score across Financial benchmarks.
ODA-Fin-RL/SFT-8B demonstrates strong performance relative to thinking models with significantly more parameters.}
  \label{main_performance_fig}

\end{figure}

\section{Introduction}
\label{section:intro}

The rapid advancement of Large Language Models (LLMs) has demonstrated remarkable capabilities across diverse tasks, including natural language understanding, reasoning, and knowledge-intensive decision making~\citep{openai2024gpt4technicalreport,qwen3technicalreport,anthropic2025claude4systemcard}. While these models show strong generalization ability in open-domain settings, their deployment in vertical domains, particularly finance, remains challenging~\citep{guo2023chatgpt, li2023large, wu2023bloomberggpt, mahdavi2025integrating, khak2025bridging}. Financial applications often involve high-stakes decision making, where errors may lead to significant economic consequences and thus require precise reasoning and reliable outputs. Financial tasks are distinctively characterized by a high density of domain-specific terminology, rigorous requirements for numerical reasoning, and a strict intolerance for factual hallucinations~\citep{tang2025financereasoning,xie2024finben,srivastava2024evaluating}. These characteristics expose fundamental limitations of current LLM training paradigms, particularly in data quality, reasoning supervision, and domain-specific alignment.

Historically, the prevailing paradigm in domain adaptation has been ``Model-Centric'', emphasizing architectural modifications or the scaling of parameters~\citep{wu2023bloomberggpt, shah-etal-2022-flang,shi2025kronos}. Yet, recent trends indicate a pivotal shift towards ``Data-Centric'' AI, acknowledging that data quality often supersedes model complexity~\citep{cai2025opendataarena, xie2023pixiu, yue2023disclawllm, phogat-etal-2024-fine}. Despite this shift, the field lacks a systematic understanding of what specifically constitutes ``high-quality'' data within the financial context.

Rather than introducing a new training objective or architecture innovation, we take a data-centric perspective and conduct a controlled empirical study on post-training data for financial LLMs. Our key premise is not that algorithms are irrelevant-indeed, robust RL optimization (e.g., GRPO~\citep{shao2024deepseekmath}) is necessary-but that in specialized vertical domains, performance is often largely determined by the quality and the difficulty/verifiability profile of post-training data. Concretely, we posit a two-stage data hierarchy: (i) during SFT, data purity and high-quality reasoning traces establish a strong instruction-following and domain-knowledge foundation; (ii) during RL, selecting hard-but-verifiable samples (i.e., tasks that are challenging yet reliably scorable by a verifier) is critical for pushing the model beyond the SFT plateau and enabling systematic capability improvements in complex financial reasoning tasks.

To test this premise, we construct two complementary datasets aligned with the two training stages. For SFT, we introduce \textbf{ODA-Fin-SFT-318k}, obtained via multi-stage distillation and verification to produce high-quality Chain-of-Thought (CoT) supervision at scale. For RL, we introduce \textbf{ODA-Fin-RL-12k}, a subset curated not only for difficulty but also for verifiability under an efficient online verifier. Importantly, all datasets are constructed entirely from existing open-source resources, without reliance on proprietary data.
Using standard SFT and RL training pipelines as controlled settings, performance improvements can be attributed primarily to data quality, enabling us to disentangle the contributions of (a) high-quality CoT distillation and (b) difficulty- and verifiability-oriented RL data selection. We further hypothesize that RL on hard-but-verifiable samples is effective because it discourages shortcut heuristics that may be reinforced by easier SFT data, and instead promotes non-trivial reasoning behaviors required for challenging financial analysis.

We validate our approach across different benchmarks spanning various tasks, such as general financial understanding, sentiment analysis, and numerical reasoning. By post-training on the Qwen3-8B base model, our ODA-Fin-SFT/RL models consistently surpass open-source state-of-the-art (SOTA) financial LLMs of comparable size, with key comparisons highlighted in Figure~\ref{main_performance_fig}. Our contributions are summarized as follows:

\begin{itemize} %[leftmargin=*] 

\item We introduce a multi-stage distillation and verification protocol that transforms raw financial Q\&A pairs into high-quality CoT data, incorporating semantic deduplication, CoT synthesis via large-scale reasoning models, and length-adaptive verification to ensure factual correctness.

\item We study verifier design and the reward precision-diversity trade-off in financial RL, showing that hard-but-verifiable sample selection, together with an appropriate answer-length constraint, is crucial for stable gains.

\item We release ODA-Fin-SFT-318k, a curated dataset from over 25 open-source repositories spanning 318K samples for SFT, along with ODA-Fin-RL-12k for reinforcement learning and the trained ODA-Fin-SFT-8B and ODA-Fin-RL-8B models to foster reproducible research in data-centric financial AI.

\end{itemize}

\section{Data Engineering}

In this section, we detail our pipeline for transforming raw, unstructured financial information into high-quality, instruction-following datasets.

\subsection{Data Source}
\label{subsec:data_source}

To ensure comprehensive coverage of the financial landscape, we first source the open-source datasets on HuggingFace and Github, aiming to cover all frequently used financial datasets, spanning sentiment analysis, financial reasoning, forecasting, and quantitative trading. After automatic and manual checking, we finally aggregate a total of 697,034 samples from over 25 distinct open-source repositories to form the raw foundation of our study (referred to as the Raw Pool). Further details are provided in Appendix~\ref{appd:dataset_details}.

\paragraph{Source Diversity.}
The collected source corpus is a composite of diverse financial intelligence. The largest contributor is \texttt{Finance-Instruct-500k}~\citep{josephgflowers2025financeinstruct}, which provides a broad base of general financial instructions. Notably, we also include recent reasoning-intensive datasets such as \texttt{DianJin-R1-Data}~\citep{zhu2025dianjin} and \texttt{Agentar-DeepFinance-100K}~\citep{zhao2025agentar}, which introduce complex chain-of-thought patterns. In total, we collected 946k raw samples.

\paragraph{Task Taxonomy.}
The task distribution reveals a significant characteristic of the current open-source financial data landscape. The vast majority of the data falls under Financial QA (roughly 80\%), followed by Sentiment Analysis. Highly specialized vertical tasks—such as Information Extraction, Financial Forecasting, Quantitative Trading, and Risk Analysis—comprise a negligible fraction of the total volume.

\subsection{Data Distillation and Filtering}
\label{subsec:data_distillation}

\begin{figure}[t]
\centering
  \includegraphics[width=\textwidth]{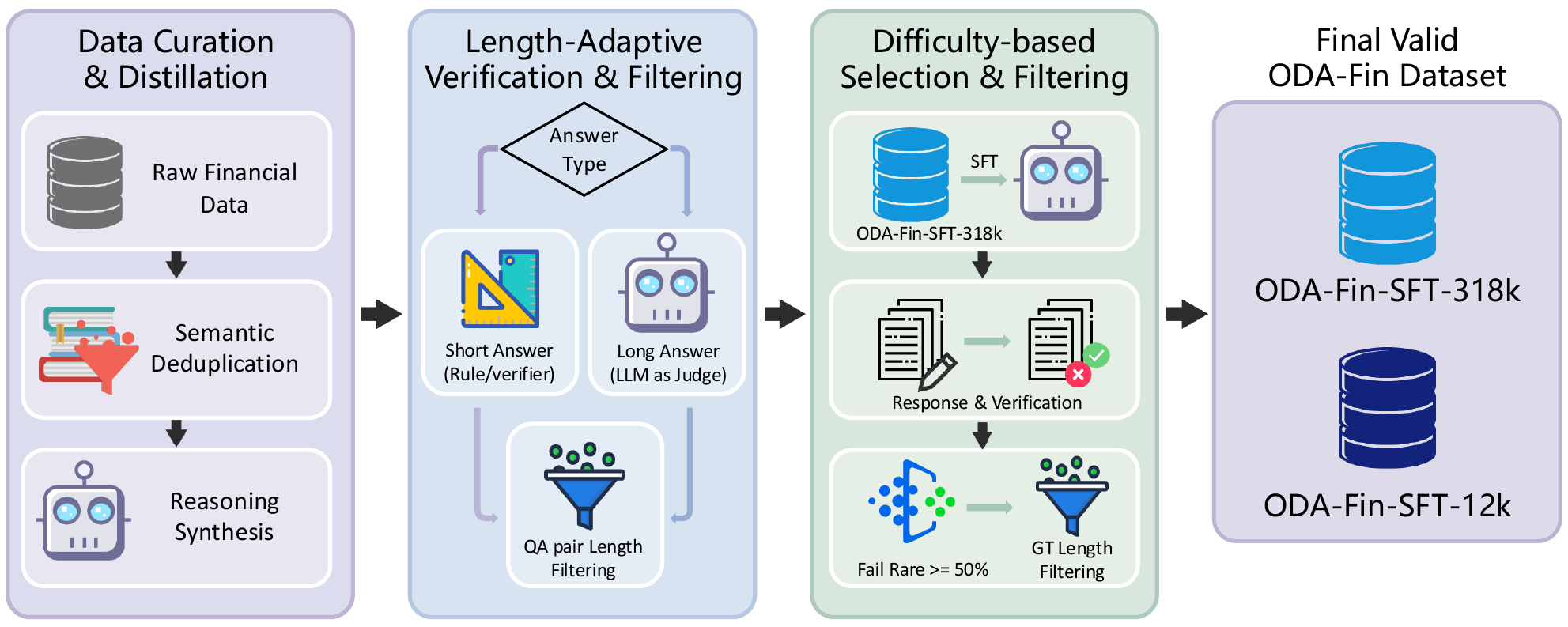}
  \caption{Data construction pipeline of ODA-Fin-SFT-318k and ODA-Fin-RL-12k.}
  \label{data_pipeline}

\end{figure}

Raw financial data often suffers from redundancy and a lack of explicit reasoning steps, the data quality is also not controlled. To construct a high-quality instruction tuning dataset, we utilize a simple yet multi-stage pipeline to process the source data, consisting of semantic deduplication, Chain-of-Thought synthesis, rigorous verification, and length filtering.

\paragraph{1) Semantic Deduplication.}
To eliminate redundancy within the collected corpus, we perform semantic deduplication with an embedding-based method. Specifically, we use \texttt{Qwen3-Embedding-8B}~\citep{qwen3embedding} to encode all samples into dense vector representations. Then, samples with cosine similarity exceeding a strict threshold are removed, ensuring diversity in the training set.

\begin{figure}[t]
\centering
  \includegraphics[width=0.8\textwidth]{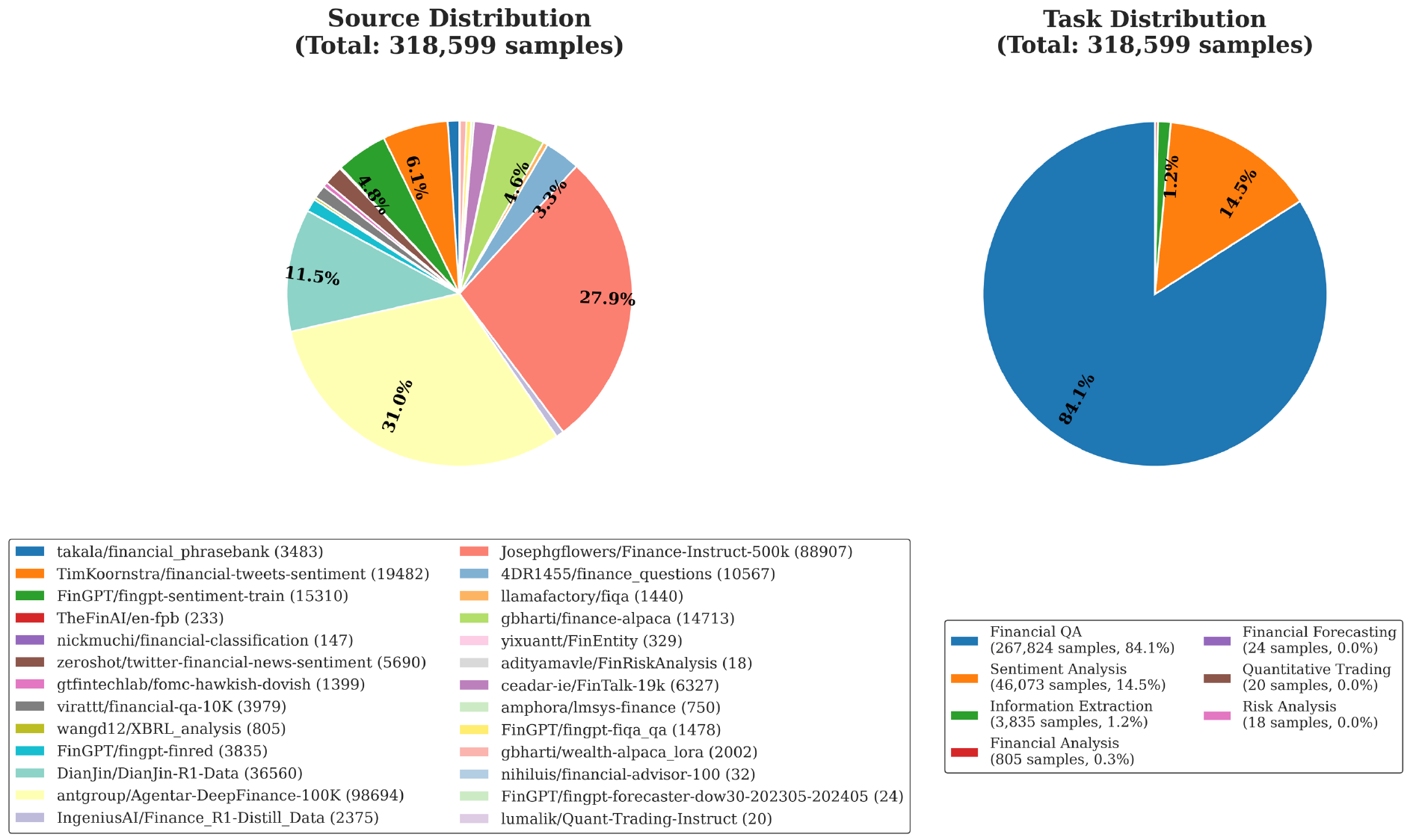}
  \caption{Data source and task distribution.}
  \label{data_dist}
\end{figure}

\paragraph{2) Reasoning Synthesis.}
A significant portion of the raw data consisted of direct Question-Answer pairs without intermediate reasoning. To upgrade these samples into reasoning-intensive data, we apply a distillation process as follows:
\begin{itemize}
    \item \textbf{CoT Generation:} For samples lacking reasoning traces, we prompt \texttt{Qwen3-235B-A22B-Thinking}\\~\citep{qwen3technicalreport} to generate a step-by-step CoT leading to the final answer. The prompt template is shown in Listing~\ref{lst:distill_prompt}.
    \item \textbf{Existing CoT Integration:} For simplicity, we retain high-quality existing CoT datasets, specifically \texttt{Agentar-DeepFinance-100K} and \texttt{DianJin-R1-Data}, which already contain rich reasoning patterns.
\end{itemize}

\paragraph{3) Length-Adaptive Verification.}
To prevent hallucinations introduced during the distillation process, we devise a length-adaptive verification protocol to filter the generated CoT data. We exclude samples where even \texttt{Qwen3-235B-A22B-Thinking} fails to produce correct answers, as these instances are likely too challenging for smaller models to learn effectively and may introduce noise into the training data.

Our verification strategy adapts to the nature of the expected output:
\begin{itemize}
    \item \textbf{Short-form Verification:} For questions requiring concise answers (e.g., classification, sentiment labeling, numerical calculation), we utilize \texttt{CompassVerifier-7B}~\citep{liu2025compassverifier}. This model acts as a reward model to strictly judge the correctness of the final prediction. We use the non-CoT version of the judge prompt provided by CompassVerifier.
    \item \textbf{Long-form Verification:} For complex queries requiring extended analysis, we employ \texttt{Qwen3-23\\5B-A22B-Instruct} as a judge. This larger model evaluates the logical coherence and factual accuracy of the generated response. The judge prompt is provided in Appendix~\ref{appd:judge_long_prompt}.
\end{itemize}

\paragraph{4) Token Length Filtering.}
To align with the context window constraints of our training infrastructure and ensure computational efficiency, we filtered out samples with extreme lengths. Any data samples exceeding 16,384 tokens were filtered out. 

After the above processing, the successfully distilled and filtered data are merged with the original high-quality CoT subsets to form our final training corpus, ODA-Fin-SFT-318k. The distribution of the processed data is illustrated in Figure~\ref{data_dist}.

\subsection{Data Selection for RL} 
For RL training, we evaluate the best-performing SFT model, ODA-Fin-SFT-8B (defined in Section~\ref{sec:main_results}), on ODA-Fin-SFT-318k to estimate the empirical difficulty of each sample. Specifically, we generate 4 answers for each data sample. Samples with a failure rate greater than 50\% are retained to prioritize challenging examples during RL training. For reliable online verification during RL, we retain only data samples whose final answer token length is less than 16. This process yields our RL training corpus, ODA-Fin-RL-12k.  The distribution of data sources and task types is shown in Figure~\ref{rl_data_dist}.

\section{Model Training}

To evaluate the data value, we use a standard training pipeline, Supervised Fine-Tuning (SFT) on distilled reasoning data and then Reinforcement Learning (RL) on difficulty-aware data.

\subsection{Supervised Fine-Tuning}
To equip the model with fundamental reasoning capabilities, we perform SFT on a strategically curated dataset.
We replace the raw data without CoT with our synthesized Distilled CoT data (generated by \texttt{qwen3-235b-thinking} and filtered via our verification pipeline), and merge it with the full set of existing high-quality CoT datasets.

The verified distilled CoT traces provide explicit intermediate reasoning steps that bridge the gap between question understanding and answer generation, which is particularly valuable for complex financial reasoning tasks involving numerical calculations and multi-step logical inference. By incorporating structured reasoning chains, the model learns not just what the correct answer is, but how to systematically arrive at it through transparent computational and logical steps. Besides, the rigorous verification process ensures that only correct reasoning paths are used for training, preventing the model from learning spurious correlations or flawed reasoning patterns that could lead to systematic errors in financial applications. The resulting training corpus thus represents a carefully curated collection of high-quality reasoning examples that balance breadth of coverage with depth of domain expertise, enabling the model to generalize effectively across diverse financial scenarios while maintaining high accuracy standards.

\subsection{Reinforcement Learning}
\label{subsec:rl}

To further enhance model performance, we conduct RL training. Unlike the offline distillation phase where we employ the computational-heavy \texttt{Qwen3-235B} as a judge, the online RL training requires efficient and low-latency reward signals. Therefore, we exclusively utilize the lightweight \texttt{CompassVerifier-7B} for real-time reward calculation. To ensure the reliability of this smaller verifier, we filter the difficulty-aware dataset to retain only samples where the ground truth is concise (e.g., classification labels, specific numbers, or short phrases). This strategy mitigates the risk of reward noise associated with verifying long-form open-ended generation, ensuring that the model receives precise feedback during the optimization process.

\paragraph{Format Reward.}
We first evaluate the structural integrity of the generated response. The model is required to enclose its reasoning process within \texttt{<think>} tags. The format reward $R_{\text{fmt}} \in [0, 1]$ is calculated as:
\begin{equation}
    R_{\text{fmt}} = 0.25 \cdot \mathbb{I}_{\text{start}} + 0.25 \cdot \mathbb{I}_{\text{end}} + 0.5 \cdot \mathbb{I}_{\text{pair}}
\end{equation}
where $\mathbb{I}_{\text{start}}$ and $\mathbb{I}_{\text{end}}$ indicate the presence of opening and closing tags, respectively, and $\mathbb{I}_{\text{pair}}$ denotes the existence of exactly one correctly ordered pair of tags.

\paragraph{Correctness Reward.}
To assess semantic correctness, we employ a hierarchical extraction strategy followed by verification. The answer is extracted by prioritizing content within \texttt{\textbackslash boxed\{\}}, falling back to regex patterns (e.g., ``The answer is...''), and defaulting to the raw response text if specific formats are absent.

The extracted answer is validated against the ground truth using either rule-based matching or a verifier model (compass-verifier-7B), which classifies the response as Correct, Incorrect, or Invalid. We define an outcome multiplier $M_{\text{outcome}}$ based on this judgment:
\begin{equation}
    M_{\text{outcome}} =
    \begin{cases}
        1.0 & \text{if Judgment} = \text{Correct} \\
        0.5 & \text{if Judgment} \in \{\text{Incorrect}, \text{Invalid}\} \text{ or extraction fails}
    \end{cases}
\end{equation}

\paragraph{Total Reward.}
The final reward $R$ is computed as the product of the format score and the outcome multiplier:
\begin{equation}
    R = R_{\text{fmt}} \times M_{\text{outcome}}
\end{equation}
This multiplicative design ensures that the model receives maximum reward only when it satisfies both formatting constraints and semantic correctness. Notably, incorrect answers still yield partial credit (scaled by 0.5) provided the reasoning format is correct, preventing sparse reward signals during early training stages.

\section{Experiments}

\subsection{Experiment Setup}

\textbf{Models and Training Configurations.}
We evaluate the data value by training on standard base models, Qwen2.5-7B-Instruct\footnote{\url{https://huggingface.co/Qwen/Qwen2.5-7B-Instruct}}~\citep{qwen2.5} and Qwen3-8B\footnote{\url{https://huggingface.co/Qwen/Qwen3-8B}}~\citep{qwen3technicalreport}.
For training configurations, we utilize the settings that are widely-adopted by other works~\citep{zhu2025dianjin, liu2025fin, duxiaoman2023xuanyuan}. Concretely, SFT is conducted on 16$\times$NVIDIA A100 GPUs, while RL training is performed on 8$\times$NVIDIA A100 GPUs. We employ the chat template consistent with Qwen3 with a maximum sequence length of 16,384 tokens. SFT Training is performed with 16 preprocessing workers and sequence packing disabled. The batch size per device is set to 1 with a gradient accumulation step of 16. We use full parameter fine-tuning with a learning rate of $1.0 \times 10^{-5}$, training for 3 epochs. The learning rate follows a cosine schedule with a warmup ratio of 0.1. We employ Group Relative Policy Optimization (GRPO)~\citep{shao2024deepseekmath} for RL training with 4 rollouts per sample. We set the rollout temperature to 0.6, top-p to 0.85, and disable top-k filtering. The training batch size is 256, with a KL divergence coefficient of 0.001 and learning rate of 1e-6.

\textbf{Evaluation Benchmarks.}
For evaluation, we assess model performance across three distinct categories of abilities with total 9 benchmarks to ensure comprehensive coverage:
\begin{itemize}
    \item \textit{General Financial Understanding.} (1) \textbf{FinEval}: designed to evaluate LLMs' financial domain knowledge and practical abilities, covering diverse financial fields such as banking, insurance, securities, finance and economics, investment research, and application security~\citep{guo-etal-2025-fineval}. (2) \textbf{Finova}: specifically designed to assess the model's agent-level financial reasoning, compliance verification, ability to handle broad financial instructions, and knowledge retrieval capabilities~\citep{zheng2025agentar}. (3) \textbf{FinanceIQ}: a challenging dataset focusing on professional financial certification exams (e.g., CPA, CFA—the gold standard in investment analysis requiring mastery of accounting, economics, and security analysis) to test deep domain expertise~\citep{duxiaoman2023financeiq}.
    \item \textit{Sentiment Analysis.} (4) \textbf{FOMC}: a sentiment classification benchmark that distinguishes hawkish versus dovish monetary policy stances in Federal Reserve communications, moving beyond traditional positive/negative sentiment analysis~\citep{shah2023trillion}. (5) \textbf{FPB}: the Financial PhraseBank dataset, consisting of sentences from financial news articles annotated with neutral, positive, or negative sentiment labels~\citep{malo2014good}. (6) \textbf{Headlines}: a collection of financial news headlines used to test the model's ability to interpret immediate market sentiment from short texts~\citep{sinha2020impactnewscommoditymarket}.
   \item \textit{Numerical Reasoning.} (7) \textbf{FinQA}: a dataset requiring models to perform complex numerical reasoning and calculations over unstructured text and tables found in financial reports~\citep{chen2021finqa}. (8) \textbf{TaTQA}: a hybrid benchmark that involves reasoning over both tabular and textual data to answer questions requiring arithmetic operations~\citep{zhu2021tat}. (9) \textbf{ConvFinQA}: a conversational dataset that requires multi-turn numerical reasoning to simulate real-world analysis of financial reports~\citep{chen2022convfinqa}.
\end{itemize}

We use the official evaluation framework for Finova\footnote{\url{https://github.com/antgroup/Finova}}. For all other benchmarks, we implement a unified inference framework backed by vLLM~\citep{kwon2023efficient}. For FOMC\footnote{\url{https://github.com/gtfintechlab/fomc-hawkish-dovish}} and FinEval\footnote{\url{https://github.com/alipay/financial_evaluation_dataset}}, we adopt the official evaluation scripts. For numerical reasoning tasks (FinQA, TaTQA, ConvFinQA), we employ CompassVerifier-7B for precise answer verification and scoring.

\textbf{Evaluation Metrics.}
Following established conventions in the literature, we use accuracy as the evaluation metric for general financial understanding benchmarks (FinEval, Finova, and FinanceIQ)~\citep{guo-etal-2025-fineval,zheng2025agentar}. For sentiment analysis tasks (FOMC, FPB, Headlines), we report weighted F1 scores, consistent with prior work in financial sentiment classification~\citep{fatemi2025comparative, inserte2023large, wang2023fingpt}. For numerical reasoning tasks (FinQA, TaTQA, ConvFinQA), we adopt accuracy following standard practices~\citep{phogat-etal-2024-fine, yang2023investlm, zhu2025dianjin, liu2025fin}.

\textbf{Baselines.}
We compare the performances of our resulted training models \textbf{ODA-Fin-SFT-8B} and \textbf{ODA-Fin-RL-8B} against five well-known open-source financial language models of corresponding sizes: (i) \textbf{FinMA-7B-full}: a comprehensive financial LLM designed to understand complex financial language and concepts~\citep{xie2023pixiu}; (ii) \textbf{Xuanyuan-6B-Chat}: a domain-specific financial LLM pre-trained from scratch on diverse financial corpora and fine-tuned with instruction tuning and RLHF to excel in financial dialogues and reasoning tasks~\citep{duxiaoman2023xuanyuan}; (iii) \textbf{Plutus-8B-Instruct}: an instruction-tuned LLM specialized for Greek financial texts~\citep{peng2025plutusbenchmarkinglargelanguage}; (iv) \textbf{Fin-R1-7B}: a 7B financial reasoning model enhanced through two-stage SFT and RL training on domain-specific CoT data to support complex reasoning in core financial business scenarios~\citep{liu2025fin}; (v) \textbf{DianJin-R1-7B}: a 7B financial LLM enhanced through two-stage SFT and GRPO training on curated financial reasoning data, designed to excel in complex financial analysis and compliance tasks~\citep{zhu2025dianjin}.

\subsection{Main Results}
\label{sec:main_results}

\begin{table}[htbp]
\centering
\small
\resizebox{\linewidth}{!}{
\begin{tabular}{l ccccccccc c}
\toprule
\multirow{3}{*}{\textbf{Model}} & 
        \multicolumn{3}{c}{\textbf{General}} & 
        \multicolumn{3}{c}{\textbf{Sentiment Analysis}} & 
        \multicolumn{3}{c}{\textbf{Numerical}} &
        \multirow{3}{*}{\textbf{AVG}} \\
        \cmidrule(lr){2-4}  \cmidrule(lr){5-7} \cmidrule(lr){8-10}
 & \textbf{FinEval} & \textbf{Finova} & \textbf{FinIQ} & \textbf{FOMC} & \textbf{FPB} & \textbf{HL} & \textbf{FinQA} & \textbf{TaTQA} & \textbf{CFQA}  \\
 & \scriptsize Acc/zh & \scriptsize Acc/zh &\scriptsize Acc/zh &\scriptsize W-F1/en &\scriptsize W-F1/en &\scriptsize W-F1/en &\scriptsize Acc/en &\scriptsize Acc/en &\scriptsize Acc/en  \\
\midrule

\multicolumn{10}{c}{\textbf{General LLMs with Reasoning}} \\  
\midrule
% gpt/gemini/kimi

\textbf{GPT-5 mini-high} &\textbf{84.6} & \textbf{67.7} & 65.9  &\textbf{66.7} &77.0 & 75.4  &\textbf{77.6} &\textbf{92.9} & \textbf{85.7}  & \textbf{77.1} \\
\textbf{Gemini 2.5 Flash-Lite} & 52.8 &51.2 &45.7  &62.5 &79.8 & 76.0  &72.0 &\underline{90.0} &79.5   & 67.7 \\

\textbf{Qwen3-4B-Thinking} &76.4 & 54.6 &72.3  &63.8 &70.2 & 75.0 &72.4 &87.6 & 80.6  & 72.5 \\

\midrule

\multicolumn{10}{c}{\textbf{General LLMs without Reasoning}} \\  
\midrule
\textbf{Llama-3.1-8B-Instruct} & 60.3 & 27.7 & 46.7 & 43.8 & 73.1 & 65.2 & 57.6 & 69.3 & 68.2 & 56.9 \\

\textbf{Qwen2.5-7B} & 26.1 & 19.0 & 36.1 & 26.5 & 59.9 & 62.0 & 52.4 & 78.3 & 69.6 & 47.8 \\

\textbf{Qwen2.5-7B-Instruct} & 78.9 & 42.0 & 65.1 & 56.4 & 78.6 & 73.2 & 64.6 & 81.5 & 72.1 & 68.0 \\

\textbf{Qwen3-8B} & 77.8 & 44.9 & 72.5 & 57.5 & 76.8 & 76.0 & 72.2 & 87.1 & 78.8 & 71.5 \\

% \textbf{Moonlight-16B-A3B} & & &  & & &  &  & &   & \\

\textbf{Qwen3-32B} &78.5 & \underline{57.4} & \textbf{80.8} &61.7 &77.3 & 76.8  & 72.4 &85.1 & \underline{82.5}  & \underline{74.7} \\

\midrule
\multicolumn{10}{c}{\textbf{Financial LLMs}} \\  
\midrule
\textbf{FinMA-7B-full} & 21.9 & - & 16.5 & 45.9 & \textbf{91.1} & \textbf{97.2} & - & - & - & - \\

\textbf{Xuanyuan-6B-Chat} & 61.1 & 12.3 & 45.5 & 43.5 & 65.5 & 18.8 & 23.4 & 49.3 & 10.9 & 36.7 \\

\textbf{Plutus-8B-Instruct} & 58.4 & 12.3 & 33.2 & 43.5 & 66.7 & 67.7 & 5.5 & 29.4 & 9.3 & 36.2 \\

\textbf{Fin-R1} & 77.2 & 38.6 & 62.2 & 50.2 & 27.1 & 73.5 & 67.7 & 80.0 & 76.2 & 61.4 \\

\textbf{Dianjin-R1-7B} & \underline{77.9} & 42.7 & \underline{76.9} & 60.8 & 76.4 & 74.5 & 67.2 & 82.0 & 74.6 & 70.3 \\

\midrule
\rowcolor[HTML]{F0FBFC}
\textbf{ODA-Fin-SFT-8B} & 76.0 & 47.8 & 72.1 & \underline{63.9} & 75.6 & 78.2 & 69.8 & 87.0 & 78.3 & 72.1 \\

\rowcolor[HTML]{F0FBFC}
\textbf{ODA-Fin-RL-8B} & 77.0 & 54.6 & 74.2 & 61.0 & \underline{83.4} & \underline{78.5} & \underline{73.3} & 89.3 & 80.4 & 74.6 \\

\bottomrule
\end{tabular}
}
\caption{Main Results: ODA-Fin-RL achieves top three performance across most benchmarks. `FinIQ', `HL' and `CFQA' refer to FinanceIQ, Headlines, and ConvFinQA benchmarks.}
\label{tab:main_results}
\end{table}

The comprehensive evaluation results are presented in Table~\ref{tab:main_results}. Our trained model, \textbf{ODA-Fin-RL-8B}, achieves 74.6\% average performance across all benchmarks-the highest among all open-source models at the 8B scale, and remarkably competitive with the significantly larger Qwen3-32B (74.7\%), while surpassing all specialized financial LLMs.

\paragraph{Overall Performance and Versatility.}
Compared to the base model Qwen3-8B, ODA-Fin-RL-8B demonstrates a significant improvement of 3.1 points on average. A particularly notable finding is that ODA-Fin-RL-8B nearly matches the performance of Qwen3-32B (74.7\%), a general-purpose model four times larger in parameter scale, underscoring the effectiveness of domain-focused data engineering over naive model scaling. When compared to existing financial LLMs, the advantage is even more pronounced. Traditional financial models such as FinMA-7B-full and Xuanyuan-6B-Chat struggle significantly with complex reasoning tasks. While recent reasoning-enhanced models like Dianjin-R1-7B show strong performance (70.3\% average), ODA-Fin-RL-8B still outperforms them by a clear margin, establishing a new SOTA among open-source financial LLMs of comparable size.

\paragraph{Numerical Reasoning Capabilities.}
The most notable improvement stems from the Numerical Reasoning benchmarks. ODA-Fin-RL-8B achieves dominant performance on FinQA (73.3\%), TaTQA (89.3\%), and ConvFinQA (80.4\%). On TaTQA, which requires hybrid reasoning over text and tables, ODA-Fin-RL-8B surpasses the strong Qwen3-8B baseline by 2.2 points, the financial competitor Dianjin-R1-7B by 7.3 points, and even the much larger Qwen3-32B by 4.2 points (89.3\% vs.\ 85.1\%). This validates that our reinforcement learning strategy, specifically the hybrid reward mechanism targeting reasoning steps, effectively enhances the model's ability to handle complex arithmetic and logic in financial reports.

\paragraph{General Financial Understanding.}
In the General Financial category, our model remains highly competitive. ODA-Fin-RL-8B achieves the best Finova performance (54.6\%) among all 8B-scale models, significantly outperforming Qwen3-8B (44.9\%) and Dianjin-R1-7B (42.7\%). While Dianjin-R1-7B leads slightly on FinEval (77.9\% vs.\ 77.0\%) and FinanceIQ (76.9\% vs.\ 74.2\%), and the larger Qwen3-32B achieves stronger general understanding scores on FinanceIQ (80.8\%), ODA-Fin-RL-8B maintains robust scores across all general understanding benchmarks, ensuring that the gains in numerical reasoning do not come at the cost of general domain knowledge.

\paragraph{Sentiment Analysis.}
For Sentiment Analysis, ODA-Fin-RL-8B demonstrates balanced performance. While specialized models like FinMA-7B-full achieve near-perfect scores on specific tasks like FPB (91.1\%) and Headlines (97.2\%), they fail to generalize to other domains. ODA-Fin-RL-8B achieves a strong 83.4\% on FPB and remains competitive on FOMC and Headlines, outperforming even the larger Qwen3-32B on both FPB (83.4\% vs.\ 77.3\%) and Headlines (78.5\% vs.\ 76.8\%), offering a more reliable solution for diverse sentiment tasks compared to models that overfit specific formats. These results suggest that further efforts are needed from a data-centric perspective to better balance task-specific data mixtures.

\paragraph{Impact of Reinforcement Learning.}
The comparison between \textbf{ODA-Fin-SFT-8B} and \textbf{ODA-Fin-RL-8B} highlights the critical role of the RL stage. The RL process yields consistent improvements across almost all metrics, raising the average score from 72.1\% to 74.6\%. The gains are particularly evident in the Finova (+6.8 points) and FinQA (+3.5 points) benchmarks, suggesting that the RL alignment successfully refines the model's instruction-following and reasoning capabilities beyond what supervised fine-tuning alone can achieve.

\subsection{Ablation Study on SFT Data Composition}
\label{sec:exp_sft}

\begin{table}[htbp]
\centering
\small
\resizebox{\linewidth}{!}{
\begin{tabular}{l c ccccccccc c}
\toprule

\multirow{3}{*}{\textbf{Model}} & 
        \multirow{3}{*}{\textbf{Data}} &
        \multicolumn{3}{c}{\textbf{General}} & 
        \multicolumn{3}{c}{\textbf{Sentiment Analysis}} & 
        \multicolumn{3}{c}{\textbf{Numerical}} &
        \multirow{3}{*}{\textbf{AVG}} \\
        \cmidrule(lr){3-5}  \cmidrule(lr){6-8} \cmidrule(lr){9-11}
        
 & & \textbf{FinEval} & \textbf{Finova} & \textbf{FinIQ} & \textbf{FOMC} & \textbf{FPB} & \textbf{HL} & \textbf{FinQA} & \textbf{TaTQA} & \textbf{CFQA}  \\

 & & \scriptsize Acc/zh & \scriptsize Acc/zh &\scriptsize Acc/zh &\scriptsize W-F1/en &\scriptsize W-F1/en &\scriptsize W-F1/en &\scriptsize Acc/en &\scriptsize Acc/en &\scriptsize Acc/en  \\
\midrule

\textbf{Qwen2.5-7B-Instruct} & - & \textbf{78.9} & 42.0 & 65.1 & 56.4 & 78.6 & 73.2 & 64.6 & 81.5 & 72.1 & 68.0 \\ \\
\midrule

\textbf{Qwen2.5-SFT-raw} & all 698K raw & 75.6 & 39.6 & 64.7 & \textbf{70.4} & \textbf{94.1} & 73.9 & 64.8 & 80.5 & 73.7 & 70.8 \\

\textbf{Qwen2.5-SFT-partial} & 192K partial cot & 73.6 & \underline{43.6} & 72.2 & 65.3 & 74.1 & 74.6 & 65.7 & 82.3 & 71.7 & 69.2 \\

\textbf{Qwen2.5-SFT-all} & all 318K cot &75.6&43.2&\textbf{72.7} &\underline{64.8}&75.3&\textbf{78.4} &68.4&86.6&73.8 & \underline{71.0} \\

\midrule

\midrule

\textbf{Qwen3-8B} & - & \underline{77.8} & 44.9 & \underline{72.5} & 57.5 & \underline{76.8} & 76.0 & \textbf{72.2} & \textbf{87.1} & \textbf{78.8} & 71.5 \\
\midrule

\textbf{Qwen3-SFT-raw} & all 697K raw &76.1&43.0&62.3 &59.0&60.9&71.9 &67.0&84.0&70.3 & 66.1 \\

\textbf{Qwen3-SFT-partial} & 192K partial cot & 74.2 & 36.8 & 70.6 & 65.7 & 73.8 & 74.4 & 66.4 & 84.0 & 75.6 & 69.1 \\

\textbf{Qwen3-SFT-mixed} & +General\&Math & 72.1 & 41.7 & 66.2 & 61.6 & 69.5 & 75.6 & 64.9 & 82.6 & 74.0 & 67.6 \\

\textbf{Qwen3-SFT-cal} & +table calculation & 70.0 & 38.9 & 63.6 & 63.1 & 70.3 & 75.8 & 63.0 & 79.8 & 69.8 & 66.0 \\

\textbf{ODA-Fin-SFT-8B} & all 318K cot & 76.0 & \textbf{47.8} & 72.1 & 63.9 & 75.6 & \underline{78.2} & \underline{69.8} & \underline{87.0} & \underline{78.3} & \textbf{72.1} \\

\bottomrule
\end{tabular}
}
\caption{SFT Data Ablation Results. "all cot" (Setting III) achieves the highest average performance. `FinIQ', `HL' and `CFQA' refer to FinanceIQ, Headlines, and ConvFinQA benchmarks.}
\label{tab:sft_ablation}
\end{table}

To identify the most effective data strategy for the SFT stage, we systematically investigate the impact of data quality, reasoning density, and domain specificity through a comparative analysis across four distinct data composition settings:
\begin{itemize}
    \item \textbf{Setting I: Raw Data Only (Baseline).} We fine-tune the model exclusively on the 697K raw dataset, which has undergone deduplication only. This configuration serves as a baseline to evaluate the model's performance when trained solely on originally collected data, thereby quantifying the value added by the distillation process.
    
    \item \textbf{Setting II: Partial CoT.} We supplement the raw dataset with a curated subset of distilled data and existing high-quality CoT data (specifically, samples from \texttt{Agentar-DeepFinance-100K} and \texttt{DianJin-R1-Data}). This experiment tests whether a modest amount of reasoning data can yield performance comparable to training on the full raw dataset.
    
    \item \textbf{Setting III: Distilled + Math + General CoT.} Hypothesizing that expert-level financial analysis requires both numerical proficiency and broad logical deduction, we explore a comprehensive data augmentation strategy. We merge our domain-specific distilled CoT data with two supplementary subsets: 5k samples of mathematical reasoning from \texttt{DeepMath-103K}\citep{he2025deepmath} to enhance calculation precision, and 5k samples of general-purpose CoT data from \texttt{Ring-Light}\citep{ringteam2025ringlitescalablereasoningc3postabilized} to strengthen logical coherence. This configuration evaluates the synergistic effect of reinforcing both calculation skills and general reasoning capabilities alongside domain knowledge.
    
    \item \textbf{Setting IV: Distilled + Table-based Calculation.} Real-world financial analysis frequently demands rigorous numerical reasoning over structured data such as balance sheets and earnings reports. We augment our distilled corpus with 545 CoT samples distilled and verified from MultiHiertt~\citep{zhao-etal-2022-multihiertt}, which focuses specifically on table-based calculation. This setting evaluates whether targeted training on tabular reasoning enhances performance beyond training on financial data alone.
\end{itemize}

Table~\ref{tab:sft_ablation} presents our ablation study results comparing these strategies across Qwen2.5-7B and Qwen3-8B architectures. From the results, we have several observations. 

\paragraph{Raw Data Hurts More as Model Capability Scales.} 
Setting I (Raw Data Only) reveals a striking architecture-dependent pattern. For Qwen2.5-7B, raw data training (\textbf{Qwen2.5-SFT-raw}) lifts the average from the base model's 68.0\% to 70.8\%, driven by strong gains in sentiment analysis (FPB: 94.1\%, FOMC: 70.4\%), suggesting that direct QA pairs still provide useful signal for less capable base models. In stark contrast, applying the identical strategy to the more capable Qwen3-8B (\textbf{Qwen3-SFT-raw}) produces a catastrophic degradation, dropping the average from 71.5\% to 66.1\%, with regressions across nearly all benchmarks (e.g., FPB: 76.8\%~$\rightarrow$~60.9\%, TaTQA: 87.1\%~$\rightarrow$~84.0\%, FinanceIQ: 72.5\%~$\rightarrow$~62.3\%). This divergence reveals a quality threshold: as base models become more capable, they have already internalized basic patterns and are disproportionately harmed by the noise and inconsistency of unstructured raw data.

\paragraph{Partial CoT Provides Inconsistent Gains.}
Introducing partial CoT (Setting II) yields mixed results across both architectures. For Qwen2.5, \textbf{Qwen2.5-SFT-partial} drops the average from 70.8\% to 69.2\% despite improvements on FinanceIQ (72.2\%) and TaTQA (82.3\%), while FPB collapses dramatically (94.1\%~$\rightarrow$~74.1\%). For Qwen3, \textbf{Qwen3-SFT-partial} recovers to 69.1\% from the raw setting's 66.1\% but remains below the Qwen3 base on aggregate performance, with inconsistent patterns including a significant drop on Finova (44.9\%~$\rightarrow$~36.8\%) despite gains on FinanceIQ (70.6\%) and numerical tasks like CFQA (75.6\%). These results consistently demonstrate that mixing small amounts of reasoning data with large raw datasets produces diluted signals and unstable cross-task performance.

\paragraph{Full Distilled CoT Achieves Consistent Best Performance.} 
Training exclusively on the complete 318K distilled CoT dataset proves decisive across both architectures. For Qwen2.5, \textbf{Qwen2.5-SFT-all} achieves the highest average among all Qwen2.5 variants at 71.0\%, with substantial gains over the raw baseline in FinanceIQ (72.7\% vs. 64.7\%), Headlines (78.4\% vs. 73.9\%), and TaTQA (86.6\% vs. 80.5\%), demonstrating that comprehensive reasoning-signal training yields consistent cross-domain improvements over both raw and partial CoT strategies. For Qwen3, \textbf{ODA-Fin-SFT-8B} achieves 72.1\%, outperforming all augmentation variants and surpassing the strong Qwen3 base on key benchmarks (Finova: 47.8\% vs. 44.9\%, Headlines: 78.2\% vs. 76.0\%, TaTQA: 87.0\% vs. 87.1\%, CFQA: 78.3\% vs. 78.8\%). Notably, the full CoT setting is the only configuration that successfully improves upon the Qwen3 base across general knowledge (Finova), sentiment analysis (Headlines), and maintains competitive performance on numerical reasoning tasks. The consistent superiority of full distilled CoT across two architectures of differing capability confirms that complete replacement with high-quality verified reasoning traces is the optimal data strategy.

Training exclusively on the complete 318K distilled CoT dataset proves decisive across both architectures. For Qwen2.5, \textbf{Qwen2.5-SFT-all} achieves the highest average among all Qwen2.5 variants at 71.0\%, with substantial gains in TaTQA (86.6\%), and Headlines (78.4\%), demonstrating that comprehensive reasoning-signal training yields consistent cross-domain improvements over both raw and partial CoT strategies. For Qwen3, \textbf{ODA-Fin-SFT-8B} achieves 72.1\%, outperforming all augmentation variants and surpassing the strong Qwen3 base on key benchmarks (Finova: 47.8\% vs. 44.9\%, Headlines: 78.2\% vs. 76.0\%). The consistent superiority of full distilled CoT across two architectures of differing capability confirms that complete replacement with high-quality verified reasoning traces is the optimal data strategy.

\paragraph{Impact of Data Augmentation.}
Both augmentation strategies (Settings III and IV) degrade performance. Adding general Math and CoT data (\textbf{Qwen3-SFT-mixed}) drops the average to 67.6\%, with regressions on FinEval (72.1\%) and Finova (41.7\%), suggesting distribution shifts that dilute financial specialization. Table-calculation augmentation (\textbf{Qwen3-SFT-cal}) results in the lowest performance (66.0\%), with the small dataset size (545 samples) causing negative transfer across benchmarks. These results confirm that exclusive reliance on high-quality, domain-specific distilled CoT data is optimal for balancing financial knowledge and numerical reasoning capabilities.

\subsection{Analysis of RL Data and Reward Designs}
\label{sec:exp_rl}

\begin{table}[htbp]
\centering
\small
\resizebox{\linewidth}{!}{
\begin{tabular}{l ccc ccccccccc c}
\toprule

\multirow{3}{*}{\textbf{Model}} & 
        \multirow{3}{*}{\textbf{Ver}} & 
        \multirow{3}{*}{\textbf{\#Tok}} &
        \multirow{3}{*}{\textbf{\#Size}} &
        \multicolumn{3}{c}{\textbf{General}} & 
        \multicolumn{3}{c}{\textbf{Sentiment Analysis}} & 
        \multicolumn{3}{c}{\textbf{Numerical}} &
        \multirow{3}{*}{\textbf{AVG}} \\
        \cmidrule(lr){5-7}  \cmidrule(lr){8-10} \cmidrule(lr){11-13}

 & & & & \textbf{FinEval} & \textbf{Finova} & \textbf{FinIQ} & \textbf{FOMC} & \textbf{FPB} & \textbf{HL} & \textbf{FinQA} & \textbf{TaTQA} & \textbf{CFQA}  \\

 & & & & \scriptsize Acc/zh & \scriptsize Acc/zh &\scriptsize Acc/zh &\scriptsize W-F1/en &\scriptsize W-F1/en &\scriptsize W-F1/en &\scriptsize Acc/en &\scriptsize Acc/en &\scriptsize Acc/en  \\
\midrule

\textbf{Qwen3-8B} & - & - & - & 77.8 & 44.9 & 72.5 & 57.5 & 76.8 & 76.0 & 72.2 & 87.1 & 78.8 & 71.5 \\

\textbf{Fin-Zero-RL} & Rule & 1 & 6k & \textbf{81.5} & 20.5 & 72.1 & 55.7 & 76.3 & 75.5 & 71.3 & 82.9 & 80.1 & 68.4 \\

\midrule
\textbf{ODA-Fin-SFT-8B} & - & - & - & 76.0 & 47.8 & 72.1 & \textbf{63.9} & 75.6 & 78.2 & 69.8 & 87.0 & 78.3 & 72.1 \\

\textbf{Fin-RL-1} & Rule & 1 & 6k & 79.0 & 46.7 & \textbf{75.8} & 63.7 & 75.1 & 78.0 & 72.1 & 88.9 & 80.2 & 73.3 \\

\textbf{Fin-RL-2} & Model & 1 & 6K & 77.3 & 56.5 & 74.5 & 59.6 & 82.6 & \textbf{78.8} & \textbf{75.0} & 89.2 & 75.0 & 74.3 \\

\textbf{Fin-RL-3} & Model & 128 & 25K & 74.7 & 45.9 & 74.0 & 58.8 & 82.1 & \textbf{78.8} & 73.5 & \textbf{89.9} & \textbf{80.4} & 73.1 \\

\textbf{ODA-Fin-RL-8B} & Model & 16 & 12K & 77.0 & \textbf{54.6} & 74.2 & 61.0 & \textbf{83.4} & 78.5 & 73.3 & 89.3 & \textbf{80.4} & \textbf{74.6} \\

\bottomrule
\end{tabular}
}

\caption{Ablation study on RL stage. `\#Tok' denotes the maximum final answer's token length constraint used for filtering RL training data, `\#Size' refers the data size. `FinIQ', `HL' and `CFQA' refer to FinanceIQ, Headlines, and ConvFinQA benchmarks.}
\label{tab:rl_ablation}
\end{table}

We then investigate the effectiveness of different design choices during the RL phase. Our experiments compare two base models: Qwen3-8B and the best-performing SFT model. For answer correctness verification, we evaluate both rule-based matching and model-based verification using \texttt{CompassVerifier-7B}. When using model-based verification, we experiment with different answer length constraints of 128, 16, and 1 tokens to study the trade-off between answer quality and generation efficiency.

Table~\ref{tab:rl_ablation} presents our ablation study on the RL stage, examining the impact of base model selection, verifier types, and answer length constraints on data filtering.

\textbf{Base Model Selection.} Comparing Qwen3-8B with our SFT model (ODA-Fin-SFT-8B) as RL base models, we observe that starting from ODA-Fin-SFT-8B consistently yields superior performance. The baseline Qwen3-8B achieves an average score of 71.5\%, while ODA-Fin-SFT-8B reaches 72.1\%. When applying RL from the Qwen3-8B base model, the resulted \textbf{Fin-Zero-RL}'s performance degrades significantly to 68.4\%, particularly on Finova (20.5\%). In contrast, RL training from ODA-Fin-SFT-8B maintains stable performance across all benchmarks, demonstrating that a strong SFT initialization is crucial for effective RL training.

\textbf{Verifier Type.} We compare rule-based and model-based (compass-verifier-7B) verification strategies. Rule-based verification (\textbf{Fin-RL-1} with Rule, 1 token) achieves 73.3\% average accuracy, showing improvements on numerical reasoning tasks (FinQA: 72.1\%, TaTQA: 88.9\%, ConvFinQA: 80.2\%) but struggling on Finova (46.7\%). Model-based verification demonstrates more robust performance, with the best configuration (ODA-Fin-RL-8B, 16 tokens) reaching 74.6\% average accuracy. The verifier model-based method with shorter final answers proves particularly effective on sentiment analysis tasks, achieving 82.6-83.4\% on FPB compared to 75.1\% with rule-based matching, alongside substantial gains on Finova (54.6-56.5\% vs. 46.7\%).

\begin{figure}[t]
\centering
  \includegraphics[width=\textwidth]{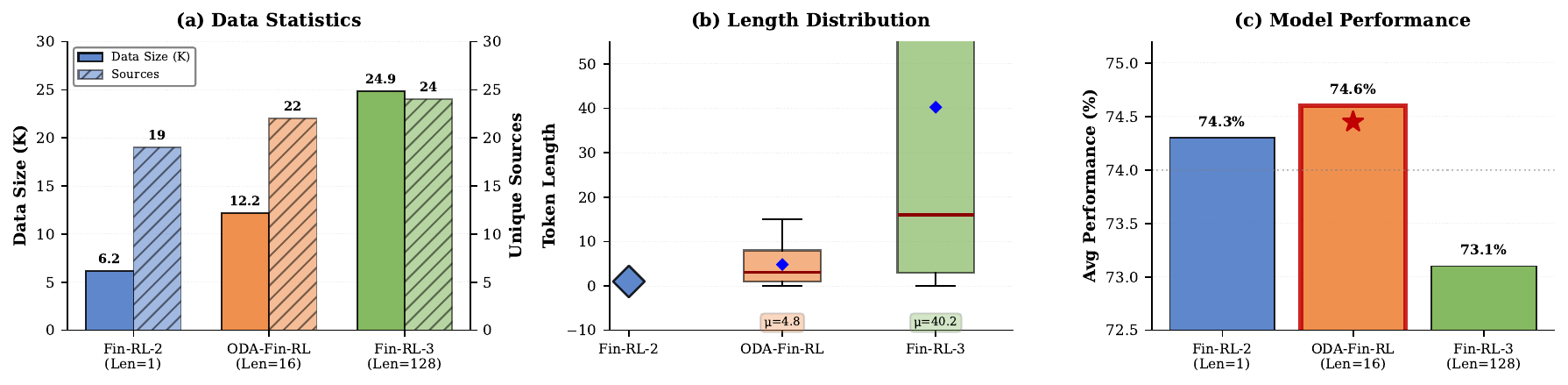}
  \caption{Ablation study on answer token length constraints. (a) Data size and source diversity; (b) Token length distribution of reference answers; (c) Average model performance across benchmarks.}
  \label{rl_ablation}
\end{figure}

\paragraph{Answer Length Constraints for Data Filtering (The Precision-Diversity Trade-off).}
A critical challenge in financial RL is balancing reward signal accuracy against task diversity. As shown in Figure~\ref{rl_ablation}, restricting reference answers to a single token (\textbf{Fin-RL-2}, 6.2K samples, 19 sources) ensures high verification precision but limits task complexity, yielding 74.3\% average performance. Conversely, relaxing the constraint to 128 tokens (\textbf{Fin-RL-3}, 24.9K samples, 24 sources) increases diversity but degrades performance to 73.1\% due to the verifier's difficulty in judging longer responses (mean $\mu=40.2$ tokens), introducing noisy rewards. Our experiments show that a 16-token constraint (ODA-Fin-RL, 12.2K samples, 22 sources) strikes the optimal balance: it retains moderate answer length ($\mu=4.8$ tokens) for reliable verification while expanding coverage to numerical reasoning and phrase extraction tasks, achieving the best overall performance of 74.6\%.

Overall, our best configuration (ODA-Fin-RL-8B with model-based verifier and 16-token constraint) achieves consistent improvements across diverse task types, validating our RL framework's effectiveness in balancing data quality and quantity for enhanced reasoning and answer accuracy.

\section{Discussion \& Insights}

In this work, we depart from the traditional "Model-Centric" paradigm to rigorously examine the primacy of data quality and training strategies within the financial domain. Our findings regarding ODA-Fin-SFT not only establish a new SOTA for open-source financial LLMs but also provide critical empirical validation for the broader "Data-Centric AI" initiative championed by OpenDataArena (ODA)~\citep{cai2025opendataarena}. We synthesize our key insights below.

\textbf{Rigorous data engineering outperforms naive scaling in vertical domains.} Our experiments reveal that domain adaptation success depends on how data is processed, not merely its volume. As shown in Section~\ref{sec:exp_sft}, mixing raw or general data degraded performance versus training on distilled, high-quality CoT data. Critically, we achieved SOTA using \textbf{only} publicly accessible sources—systematic distillation of open repositories outperformed models trained on vastly larger proprietary corpora. This demonstrates that dataset value, as benchmarked by ODA, lies in rigorous quality engineering rather than exclusivity or scale.

\textbf{Reinforcement Learning transforms financial LLMs from knowledge retrievers into reasoning experts.} Our results demonstrate that financial intelligence requires mastering domain-specific logic rather than mere knowledge accumulation. The ODA-Fin-RL-8B model's TaTQA performance (89.3\%) validates this: explicit CoT generation paired with difficulty-oriented RL (targeting samples with $\geq 50\%$ failure rate) proved essential to transcend SFT's imitation learning ceiling. Critically, our reward mechanism with a 16-token verifier constraint strikes an optimal balance—shorter constraints limit diversity while longer ones introduce noise. This targeted approach prioritizes behavioral cloning of expert reasoning processes over naive corpus expansion.

\textbf{Indiscriminate data augmentation induces negative transfer due to domain distribution shifts.}
Contrary to the intuition that general mathematical skills transfer directly to finance, we observed that mixing general mathematical data (\textit{DeepMath}) and general CoT data (\textit{Ring-Light}) with our financial corpus (Setting III) harmed performance. This negative transfer suggests that financial reasoning relies on specific conventions (e.g., accounting standards, implicit context in earnings calls) that are distinct from pure mathematical logic. This reinforces the ODA perspective on "Data Genealogy": understanding the lineage and composition of datasets is crucial. Indiscriminate mixing of high-quality but out-of-domain data can dilute the specialized capabilities required for vertical tasks, emphasizing the need for domain-aligned data curation.

\section{Conclusion}
\label{section:conclusion}

We validate the critical role of data engineering in financial LLM development by introducing ODA-Fin-SFT-318k and ODA-Fin-RL-12k. Our comprehensive empirical study demonstrates that data quality, difficulty, and verifiability—rather than volume or algorithmic novelty—determine model performance ceilings in vertical domains. Through controlled SFT and RL experiments, we show that high-quality CoT distillation establishes a robust instruction-following foundation, while our difficulty- and verifiability-aware RL strategy optimizes the trade-off between reward precision and task diversity to drive superior generalization. Our ODA-Fin-RL-8B model achieves SOTA results across nine benchmarks, with particularly strong improvements in numerical reasoning tasks. We release our datasets and trained models to advance the paradigm shift from model-centric to data-centric AI in financial intelligence research.

% \section{Acknowledgments}
% \label{section:ack}

\clearpage
\newpage
\bibliographystyle{plainnat}
\setcitestyle{numbers}
\bibliography{paper}

\clearpage
\newpage
\beginappendix

\section{Dataset Details}
\label{appd:dataset_details}
Table~\ref{tab:financial_datasets} provides an overview of the financial instruction datasets utilized in this study, including their sources and Hugging Face download counts as of February 2026.

\begingroup
\footnotesize
\setlength{\tabcolsep}{2.5pt}
\begin{longtable}{
  @{}>{\raggedright\arraybackslash}p{0.22\textwidth}
  @{\hspace{2pt}}>{\centering\arraybackslash}p{0.06\textwidth}
  @{\hspace{2pt}}>{\centering\arraybackslash}p{0.06\textwidth}
  @{\hspace{2pt}}>{\centering\arraybackslash}p{0.06\textwidth}
  @{\hspace{2pt}}>{\centering\arraybackslash}p{0.06\textwidth}
  @{\hspace{2pt}}>{\raggedright\arraybackslash}p{0.28\textwidth}
  @{\hspace{2pt}}>{\centering\arraybackslash}p{0.08\textwidth}
  @{\hspace{2pt}}>{\centering\arraybackslash}p{0.06\textwidth}@{}
}
\caption{Overview of financial instruction datasets used in this study. Download counts are from Hugging Face as of February 2026.}
\label{tab:financial_datasets} \\
\toprule
\textbf{Dataset} & \textbf{Year} & \textbf{Size} & \textbf{Lang} & \textbf{CoT} & \textbf{Description} & \textbf{DLs} & \textbf{Link} \\
\midrule
\endfirsthead

\multicolumn{8}{l}{\tablename\ \thetable\ -- \textit{Continued from previous page}}\\
\toprule
\textbf{Dataset} & \textbf{Year} & \textbf{Size} & \textbf{Lang} & \textbf{CoT} & \textbf{Description} & \textbf{DLs} & \textbf{Link} \\
\midrule
\endhead

\midrule \multicolumn{8}{r}{\textit{Continued on next page}}\\
\endfoot
\bottomrule
\endlastfoot

takala/financial\_phrasebank~\citep{malo2014good} & 2013 & 5K & EN & \ding{55} & Financial news sentiment classification & 71K & \href{https://huggingface.co/datasets/takala/financial_phrasebank}{\faHuggingFace} \\
\midrule

TimKoornstra/financial-tweets-sentiment & 2024 & 38K & EN & \ding{55} & Twitter financial sentiment analysis & 104 & \href{https://huggingface.co/datasets/TimKoornstra/financial-tweets-sentiment}{\faHuggingFace} \\
\midrule

FinGPT/fingpt-sentiment-train & 2024 & 77K & EN & \ding{55} & Financial sentiment training data & 302 & \href{https://huggingface.co/datasets/FinGPT/fingpt-sentiment-train}{\faHuggingFace} \\
\midrule

TheFinAI/en-fpb & 2024 & 3k & EN & \ding{55} & English financial phrasebank & 107 & \href{https://huggingface.co/datasets/TheFinAI/en-fpb}{\faHuggingFace} \\
\midrule

nickmuchi/financial-classification & 2022 & 5k & EN & \ding{55} & Financial text classification & 129 & \href{https://huggingface.co/datasets/nickmuchi/financial-classification}{\faHuggingFace} \\
\midrule

zeroshot/twitter-financial-news-sentiment & 2023 & 9.5K & EN & \ding{55} & Twitter financial news sentiment & 4.2K & \href{https://huggingface.co/datasets/zeroshot/twitter-financial-news-sentiment}{\faHuggingFace} \\
\midrule

gtfintechlab/fomc-hawkish-dovish~\citep{shah2023trillion} & 2023 & 25K & EN & \ding{55} & FOMC statement sentiment analysis & 1.4K & \href{https://github.com/gtfintechlab/fomc-hawkish-dovish}{\faGithub} \\
\midrule

virattt/financial-qa-10K & 2024 & 4K & EN & \ding{55} & Financial QA pairs & 344 & \href{https://huggingface.co/datasets/virattt/financial-qa-10K}{\faHuggingFace} \\
\midrule

wangd12/XBRL\_analysis & 2025 & 805 & EN & \ding{55} & XBRL financial statement analysis & 69 & \href{https://huggingface.co/datasets/wangd12/XBRL_analysis}{\faHuggingFace} \\
\midrule

FinGPT/fingpt-finred & 2024 & 3.8K & EN & \ding{55} & Financial relation extraction and classification & 35 & \href{https://huggingface.co/datasets/FinGPT/fingpt-finred}{\faHuggingFace} \\
\midrule

DianJin/DianJin-R1-Data~\citep{zhu2025dianjin} & 2025 & 36.6K & ZH & \ding{51} & Chinese financial reasoning data & 27 & \href{https://huggingface.co/datasets/DianJin/DianJin-R1-Data}{\faHuggingFace} \\
\midrule

antgroup/Agentar-DeepFinance-100K~\citep{zhao2025agentar} & 2025 & 99.1K & EN & \ding{55} & Agent-based financial reasoning & 36K & \href{https://huggingface.co/datasets/antgroup/Agentar-DeepFinance-100K}{\faHuggingFace} \\
\midrule

IngeniusAI/Finance\_R1-Distill\_Data & 2025 & 500K & EN & \ding{51} & Distilled financial reasoning data & 18K & \href{https://huggingface.co/datasets/IngeniusAI/Finance_R1-Distill_Data}{\faModelscope} \\
\midrule

Josephgflowers/Finance-Instruct-500k & 2025 & 518K & EN & \ding{55} & Large-scale financial instruction dataset & 14K & \href{https://huggingface.co/datasets/Josephgflowers/Finance-Instruct-500k}{\faHuggingFace} \\
\midrule

4DR1455/finance\_questions & 2024 & 54K & EN & \ding{55} & General financial questions & 11 & \href{https://huggingface.co/datasets/4DR1455/finance_questions}{\faHuggingFace} \\
\midrule

llamafactory/fiqa & 2024 & 5.5K & EN & \ding{55} & Curated FiQA dataset & 93 & \href{https://huggingface.co/datasets/llamafactory/fiqa}{\faHuggingFace} \\
\midrule

gbharti/finance-alpaca & 2023 & 68.9K & EN & \ding{55} & General finance instruction dataset & 947 & \href{https://huggingface.co/datasets/gbharti/finance-alpaca}{\faHuggingFace} \\
\midrule

yixuantt/FinEntity & 2023 & 979 & EN & \ding{55} & Financial entity recognition & 72 & \href{https://huggingface.co/datasets/yixuantt/FinEntity}{\faHuggingFace} \\
\midrule

adityamavle/FinRiskAnalysis & 2024 & 99 & EN & \ding{55} & Financial risk assessment analysis & 6 & \href{https://huggingface.co/datasets/adityamavle/FinRiskAnalysis}{\faHuggingFace} \\
\midrule

ceadar-ie/FinTalk-19k & 2024 & 19.1K & EN & \ding{55} & Financial conversation dataset & 34 & \href{https://huggingface.co/datasets/ceadar-ie/FinTalk-19k}{\faHuggingFace} \\
\midrule

amphora/lmsys-finance & 2024 & 735 & EN & \ding{55} & LMSYS financial conversations & 13 & \href{https://huggingface.co/datasets/amphora/lmsys-finance}{\faHuggingFace} \\
\midrule

FinGPT/fingpt-fiqa\_qa & 2024 & 17.1K & EN & \ding{55} & Financial QA from FiQA dataset & 124 & \href{https://huggingface.co/datasets/FinGPT/fingpt-fiqa_qa}{\faHuggingFace} \\
\midrule

gbharti/wealth-alpaca\_lora & 2023 & 44.3K & EN & \ding{55} & Instruction-following financial data & 75 & \href{https://huggingface.co/datasets/gbharti/wealth-alpaca_lora}{\faHuggingFace} \\
\midrule

nihiluis/financial-advisor-100 & -- & 100 & EN & \ding{55} & Financial advisor conversations & 59 & -- \\
\midrule

FinGPT/fingpt-forecaster-dow30 & 2024 & 1.23 & EN & \ding{55} & Stock forecasting instruction data & 164 & \href{https://huggingface.co/datasets/FinGPT/fingpt-forecaster-dow30-202305-202405}{\faHuggingFace} \\
\midrule

lumalik/Quant-Trading-Instruct & 2024 & 386 & EN & \ding{51} & Quantitative trading instructions & 39 & \href{https://huggingface.co/datasets/lumalik/Quant-Trading-Instruct}{\faHuggingFace} \\

\end{longtable}
\endgroup

\begin{figure}[htbp]
\centering
  \includegraphics[width=\textwidth]{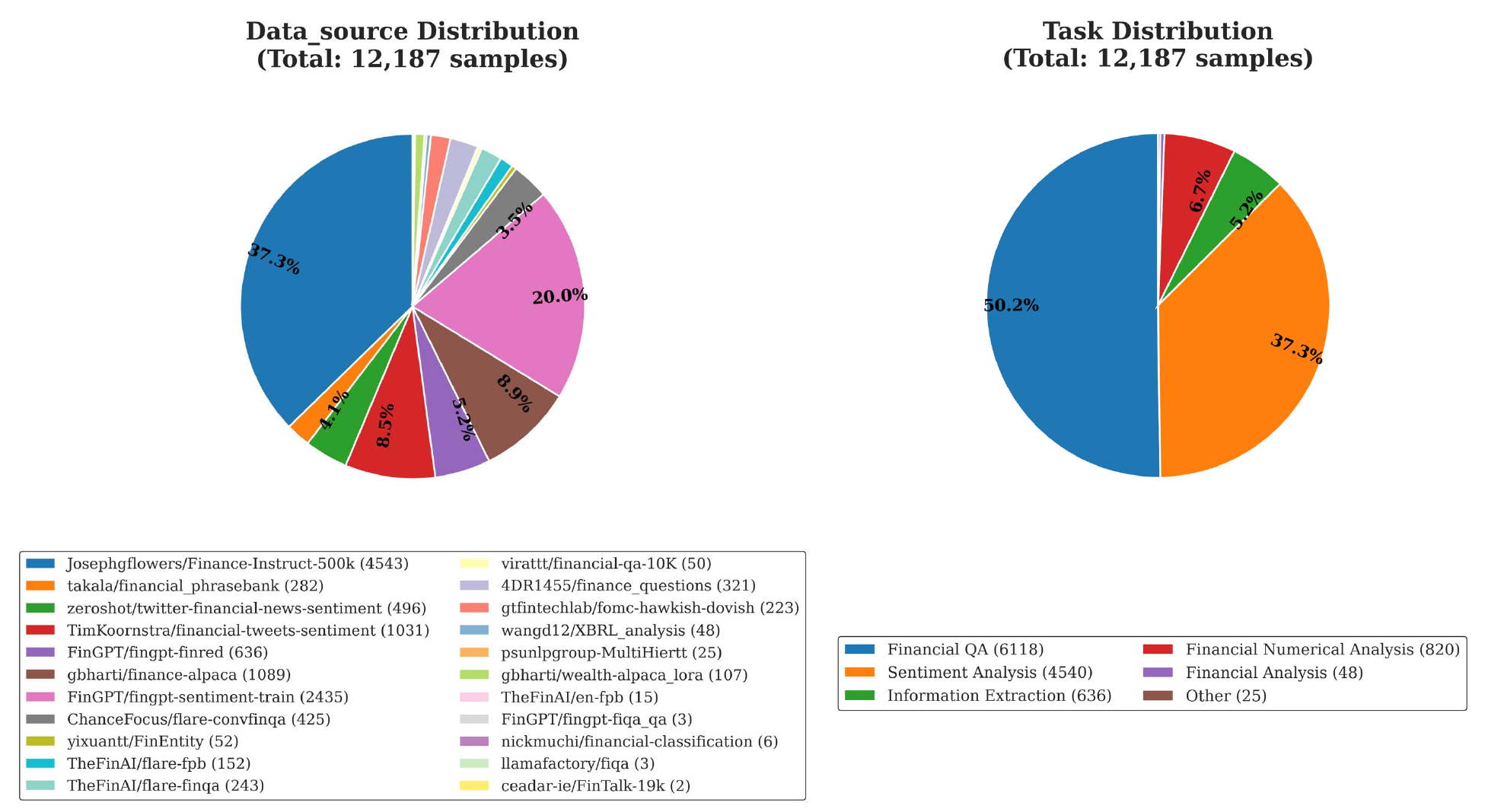}
  \caption{ODA-Fin-RL-12k data source and task disstribution.}
  \label{rl_data_dist}

\end{figure}

Figure~\ref{rl_data_dist} illustrates the specific data source composition and task category distribution for the ODA-Fin-RL-12k dataset.

\section{Prompts}
\label{appd:judge_long_prompt}

Listing~\ref{lst:distill_prompt} presents the prompt for CoT generation using Qwen3-235-A22B-Thinking. Listing~\ref{lst:judge_long_prompt} presents the specific prompt template used for model-based verification of long-form answers.

\begin{lstlisting}[caption={Prompt for CoT generation using Qwen3-235-A22B-Thinking.}, label={lst:distill_prompt}]
You are an expert in financial tasks. Reason step by step and place your final answer within \boxed{}.

Question: {question}
\end{lstlisting}

\begin{lstlisting}[caption={Prompt for evaluating correctness in long-form answers.}, label={lst:judge_long_prompt}]
You are a professional financial analysis evaluator. Your task is to assess whether a student's answer is semantically consistent with the reference answer in the context of finance and accounting.

## Evaluation Criteria:

1. **Core Financial Concepts**: The student's answer demonstrates understanding of key financial principles, metrics, and terminology (e.g., revenue, EBITDA, cash flow, ROI, NPV, financial ratios, accounting standards).

2. **Numerical Accuracy**: When financial figures, percentages, ratios, or calculations are involved, they must be accurate or reasonably close to the reference answer.

3. **Technical Completeness**: The answer includes essential financial information points such as:
   - Key financial metrics and indicators
   - Relevant time periods and fiscal years
   - Important business segments or divisions
   - Material financial events or transactions

4. **Logical Reasoning**: The financial analysis, reasoning process, and conclusions are sound and well-supported.

5. **Acceptable Variations**: Allow for:
   - Different wording or phrasing while maintaining meaning
   - Alternative but equivalent financial terminology
   - Different presentation formats (e.g., percentages vs. decimals)
   - Minor rounding differences in numerical values

## Question:
{question}

## Reference Answer:
{reference_answer}

## Student Answer:
{student_answer}

## Instructions:
Please provide your judgment directly:
- If the answers are semantically consistent (considering financial context and terminology), output: **Yes**
- If the answers are semantically inconsistent or contain material errors, output: **No**

Judgment:
\end{lstlisting}

\end{document}